\documentclass[letterpaper]{article} 
\usepackage{aaai2026}
\usepackage{times}  
\usepackage{helvet}  
\usepackage{courier}  
\usepackage[hyphens]{url}  
\usepackage{graphicx} 
\usepackage{natbib}  
\usepackage{caption} 
\usepackage{algorithm}
\usepackage{algorithmic}
\usepackage{amsmath}
\usepackage{amssymb}
\usepackage{enumitem}
\usepackage{booktabs}
\usepackage{multirow}
\usepackage{array}
\usepackage{xcolor}
\usepackage{newfloat}
\usepackage{listings}

\usepackage[most]{tcolorbox}

\DeclareCaptionStyle{ruled}{labelfont=normalfont,labelsep=colon,strut=off} 
\floatstyle{ruled}
\newfloat{listing}{tb}{lst}{}
\floatname{listing}{Listing}
\title{HieroAction: Hierarchically Guided VLM for Fine-Grained Action Analysis}

\author {
    Junhao Wu\textsuperscript{\rm 2}, 
    Xiuer Gu\textsuperscript{\rm 4},
    Zhiying Li\textsuperscript{\rm 7},
    Yeying Jin\textsuperscript{\rm 3},
    Yunfeng Diao\textsuperscript{\rm 9},
    Zhiyu Li\textsuperscript{\rm 5},
    Zhenbo Song\textsuperscript{\rm 8},
    Xiaomei Zhang\textsuperscript{\rm 6},
    Zhaoxin Fan\textsuperscript{\rm 1}\thanks{Corresponding author.}
}
\affiliations {
    
    \textsuperscript{\rm 1}Beijing Advanced Innovation Center for Future Blockchain and Privacy Computing, School of Artificial Intelligence, Beihang University; 
    \textsuperscript{\rm 2}Psyche AI Inc; 
    \textsuperscript{\rm 3}National University of Singapore; 
    \textsuperscript{\rm 4}The University of Texas at Austin; 
    \textsuperscript{\rm 5}MemTensor (Shanghai) Technology Co., Ltd.; 
    \textsuperscript{\rm 6}Institute of Automation, Chinese Academy of Sciences; 
    \textsuperscript{\rm 7}Jinan University; 
    \textsuperscript{\rm 8}Nanjing University of Science and Technology; 
    \textsuperscript{\rm 9}Hefei University of Technology; \\
    wujunhao333@gmail.com, zhaoxinf@buaa.edu.cn 
}

\usepackage{bibentry}

\begin{document}

\maketitle

\begin{abstract}
Evaluating human actions with clear and detailed feedback is important in areas such as sports, healthcare, and robotics, where decisions rely not only on final outcomes but also on interpretable reasoning. However, most existing methods provide only a final score without explanation or detailed analysis, limiting their practical applicability. To address this, we introduce HieroAction, a vision-language model that delivers accurate and structured assessments of human actions. HieroAction builds on two key ideas: (1) Stepwise Action Reasoning, a tailored chain-of-thought process designed specifically for action assessment, which guides the model to evaluate actions step by step—from overall recognition, through sub-action analysis, to final scoring—thus enhancing interpretability and structured understanding; and (2) Hierarchical Policy Learning, a reinforcement learning strategy that enables the model to learn fine-grained sub-action dynamics and align them with high-level action quality, thereby improving scoring precision. The reasoning pathway structures the evaluation process, while policy learning refines each stage through reward-based optimization. Their integration ensures accurate and interpretable assessments, as demonstrated by superior performance across multiple benchmark datasets. Code will be released upon acceptance.

\end{abstract}

\section{Introduction}
Evaluating human actions is essential in various fields, including sports, rehabilitation, and intelligent training systems~\cite{zhou2024comprehensivesurveyactionquality,dave2025finepseudo}. In sports, action analysis ensures fair and objective judging, supports athlete development, and helps prevent injuries by identifying technical issues. In rehabilitation and healthcare, it provides objective tracking of patient progress and enables customized recovery plans. Intelligent training systems further benefit from accurate action evaluation by offering practical feedback and supporting user improvement, even in the absence of a human coach~\cite{bianchi2025patsproficiencyawaretemporalsampling}. Therefore, reliable action assessment is valuable not only for elite athletes, but also for the broader population aiming to enhance performance or maintain physical well-being.

\begin{figure}[t]
  \hfill
  \includegraphics[width=0.48\textwidth]{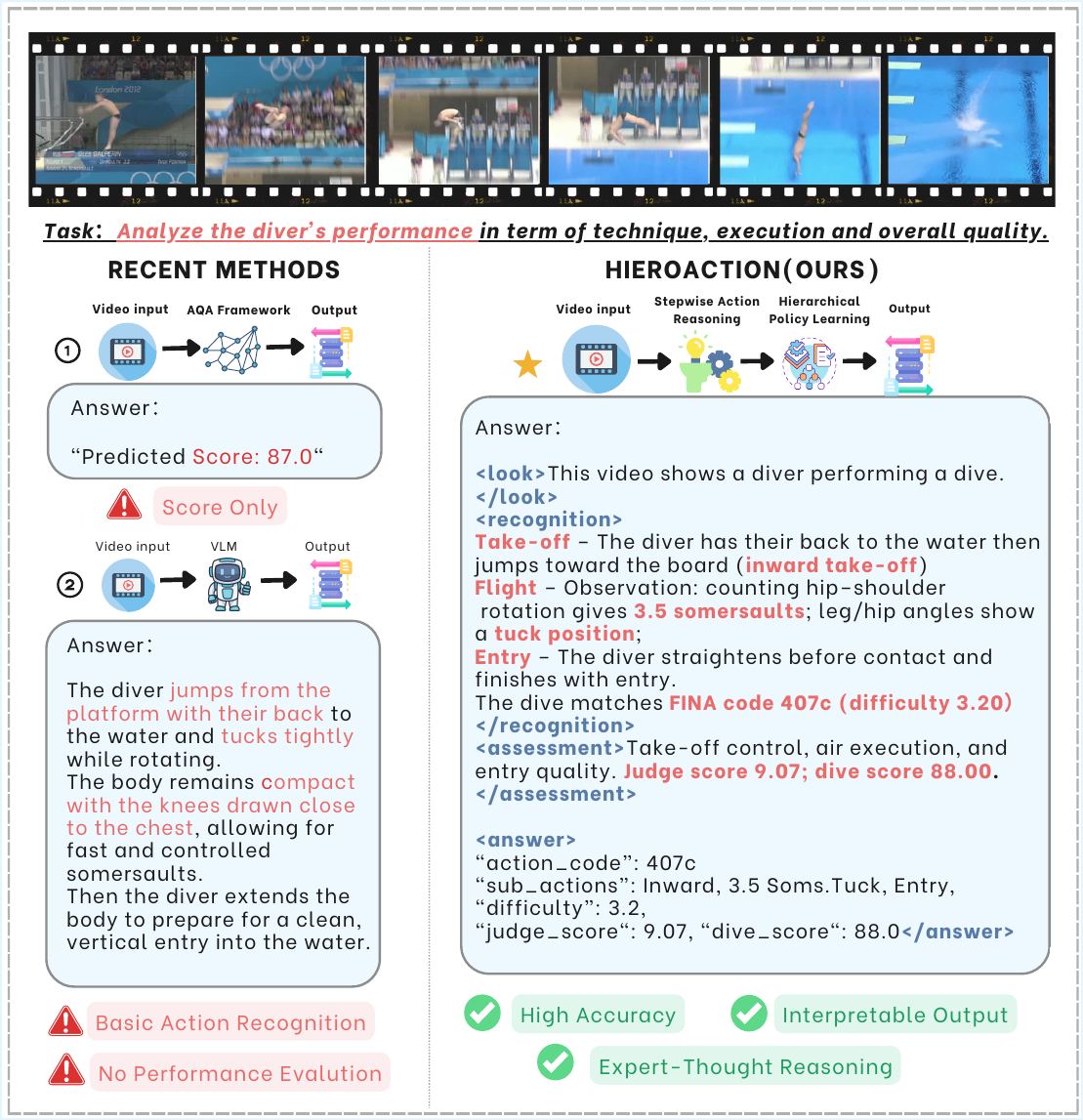}
  \caption{Motivation of HieroAction: While most existing methods offer only final scores and basic action recognition, HieroAction introduces a hierarchical framework that integrates technical standards and expert reasoning to provide precise, interpretable, and comprehensive performance assessments.}
  \label{fig:analysis example}
\end{figure}

Despite its importance, most current approaches for action analysis remain limited in scope and granularity. Traditional methods mainly focus on recognizing coarse actions or producing a single overall score for a sequence~\cite{zhou2024cofinal,xu2025visionlanguageaqa}. While such methods perform well on standard benchmarks, they often overlook the complexity and diversity of real human movements. Some recent efforts attempt to move beyond this by identifying basic sub-actions or segmenting actions into rough phases~\cite{xu2025visionlanguageaqa,reilly2025llavidallargelanguagevision}. However, these approaches still fall short of the comprehensive and fine-grained analysis expected in professional judging scenarios. Most existing techniques lack the ability to model fine temporal structures, local variations, and domain-specific technical cues~\cite{rudin2019interpretable,han2025finecausalcausalbasedframeworkinterpretable,jiang2025domainadaptationvlmsoccer}, which are crucial for high-quality evaluation. Achieving expert-level, interpretable, and fine-grained action analysis and scoring remains a challenging and open research direction.

To address the limitation, we propose \textbf{HieroAction}, a novel vision-language framework designed for precise and interpretable motion analysis, as illustrated in Fig.~\ref{fig:analysis example}. HieroAction addresses the limitations of direct score regression by decomposing motion evaluation into a structured, hierarchical process. Specifically, the framework extracts salient motion cues, applies temporal reasoning, and utilizes expert-informed prompts to produce stepwise, interpretable assessments. This hierarchical strategy enables the model to capture fine-grained execution details, segment complex actions into meaningful phases, and localize performance errors, leading to more transparent and reliable motion evaluation.

At the heart of HieroAction is the Stepwise Action Reasoning module, which adopts a tailored chain-of-thought process to structure evaluation into four stages: global action recognition, sub-action decomposition, multi-dimensional assessment, and structured output. This stepwise reasoning enables the model to progressively parse complex motions and evaluate performance in a manner that aligns with expert judgment. To further enhance fine-grained understanding and robust reasoning, we introduce Hierarchical Policy Learning with a multi-level reward structure, guiding the model from local perception to global evaluation. This approach reduces reliance on language priors, encourages visual-centric reasoning, and improves generalization across diverse action analysis tasks. 

To this end, we also propose an evaluation protocol that jointly assesses predictive accuracy and interpretability, enabling a systematic validation of our framework.  Experimental results on the FineDive, FineFS, and LOGO datasets show that our framework achieves state-of-the-art performance, highlighting its effectiveness and generalizability.  Our contributio can be summarized ass:
\begin{itemize}
  \item Conceptually, we propose \textit{HieroAction}, a vision-language framework for fine-grained action assessment. It approaches evaluation as a step-by-step reasoning process, delivering precise and interpretable results consistent with expert judgment.

  \item Technically, we design a unified architecture with two core components. \textit{Stepwise Action Reasoning} guides the model from global recognition to sub-action analysis, while \textit{Hierarchical Policy Learning} uses multi-scale rewards to refine local details and maintain overall performance.

  \item Empirically, we introduce a \textit{Multi-Dimensional Evaluation Protocol} that evaluates recognition accuracy, score reliability, and textual feedback quality. Our experiments on FineDive, FineFS, and LOGO demonstrate consistent improvements over prior methods.
\end{itemize}

\begin{figure*}[htbp]
  \centering
  \includegraphics[width=\textwidth]{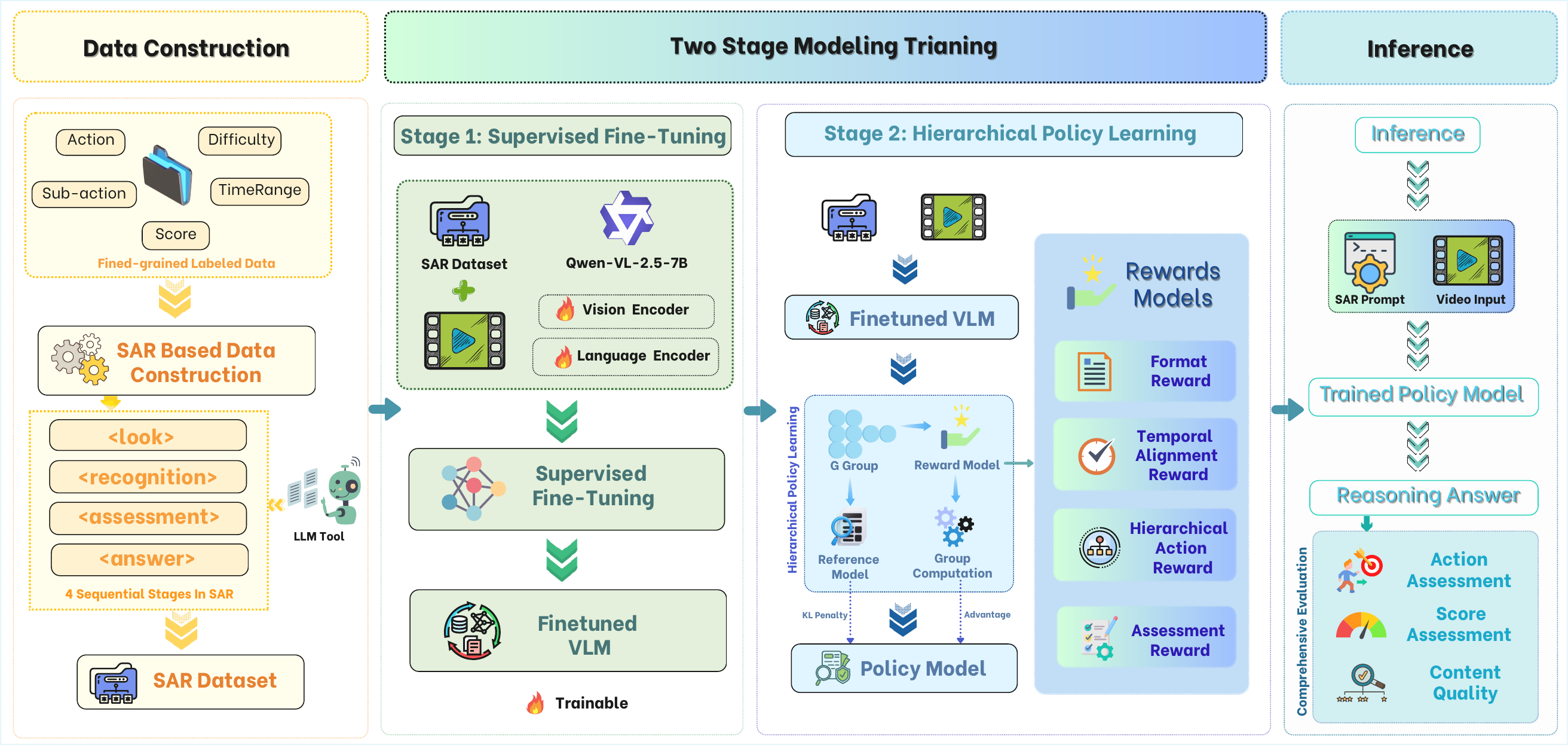}
  \caption{
Overview of the proposed \textbf{HieroAction} framework. Given a video and its fine-grained annotations, the Stepwise Action Reasoning (SAR) module guides the construction of structured and interpretable reasoning data. This data is first used to fine-tune the Qwen-VL-2.5-7B model, enabling it to learn domain-specific knowledge related to action understanding. In the second stage, Hierarchical Policy Learning (HPL) is applied to further refine sub-action analysis and scoring details through multi-level reward optimization. The final model produces accurate, robust, and interpretable results for action analysis. Comprehensive evaluation is conducted across three aspects: action recognition, score prediction, and content quality.
}
  \label{fig:framework}
\end{figure*}

\section{Related Work}

\subsection{VLMs for Human Action Understanding}

Vision-language models (VLMs) are gaining attention in human action understanding for their ability to combine visual and textual information into interpretable reasoning. Early methods relied on handcrafted features and regression models~\cite{pirsiavash2014assessing}, while recent approaches like PHI~\cite{zhou2025phibridgingdomainshift} and FineCausal~\cite{han2025finecausalcausalbasedframeworkinterpretable} introduce temporal modeling and causal reasoning for improved interpretability. VLMs address remaining limitations by using textual supervision and enhancing visual-language alignment. ChatVLA~\cite{zhou2025chatvlaunifiedmultimodalunderstanding} applies expert prompts and staged alignment, and MotionLLM~\cite{chen2024motionllm} encodes motion as symbolic tokens. Despite generalizing well, these models still rely on coarse analysis and struggle with temporal details, local variations, and domain-specific cues. To address this, we propose \textit{HieroAction}, which uses stepwise action reasoning and hierarchical policy learning to support structured and fine-grained assessment.

\subsection{Chain-of-Thought Reasoning for Structured Action Analysis}

Chain-of-Thought (CoT) reasoning enables language models to solve complex tasks through step-by-step thinking. CoT prompting~\cite{wei2022chain} improves performance by inserting intermediate steps, while Zero-Shot CoT~\cite{kojima2022zero} removes the need for demonstrations using minimal prompts. In multimodal contexts, CoT has been extended to vision-language models. Multimodal CoT~\cite{zhang2023multimodal} separates rationale generation from answer selection, Visual CoT~\cite{shao2024visualcotadvancingmultimodal} grounds reasoning in image regions, and VCTP~\cite{chen2024vcot} incorporates visual verification to reduce hallucinations. However, these methods mainly handle static inputs and overlook temporal structure, expert evaluation, and execution-level details. To address this, we propose stepwise action reasoning, which breaks actions into sequential units to capture sub-action boundaries, execution quality, and scoring logic. This helps the model align motion features with expert judgment, leading to more interpretable and reliable assessments.

\subsection{Reinforcement Learning for Fine-Grained Video Understanding}

Reinforcement learning (RL) offers a promising framework to model fine-grained video understanding as a sequential decision-making process with delayed rewards. Recent methods focus on integrating structured reasoning and multi-scale supervision. DIP-R1~\cite{park2025dipr1deepinspectionperception} and Video-R1~\cite{feng2025videor1reinforcingvideoreasoning} apply GRPO with spatial-temporal rewards to enhance temporal coherence and sensitivity. UniVG-R1~\cite{bai2025univgr1reasoningguideduniversal} combines GRPO with Chain-of-Thought supervision to align model reasoning with human logic, while VerIPO~\cite{li2025veripocultivatinglongreasoning} introduces GRPO-based feedback loops for long-horizon consistency. These approaches demonstrate the value of guided policies and hierarchical rewards for reasoning-aware video analysis. However, most still focus on saliency or global coherence, lacking phase-level cues and final scoring logic critical for action quality assessment. To address this, we propose a \textit{Hierarchical Policy Guidance} strategy under the GRPO framework. Our method uses multi-scale rewards to jointly optimize sub-action localization and high-level evaluation of action type and quality, bridging perception with expert reasoning and improving interpretability and assessment accuracy.

\section{Method}
\label{sec:method}

\subsection{Overview}
\label{sec:method_overview}

Given a video sequence $\mathcal{V} = \{I_1, I_2, \ldots, I_T\}$ depicting a complex human action, together with a task-specific prompt $q$, our objective is to learn a mapping
\begin{equation}
    \mathcal{F} : (\mathcal{V}, q) \rightarrow \mathcal{A}
\end{equation}
where $\mathcal{A} = \{(a_1, s_1, e_1), \ldots, (a_K, s_K, e_K)\}$ denotes a stepwise, interpretable assessment. Here, $a_k$ represents the $k$-th action or sub-action, $s_k$ is the expert-aligned score, and $e_k$ is the corresponding explanation for each step.

\begin{figure}[t]
  \hfill
  \includegraphics[width=0.45\textwidth]{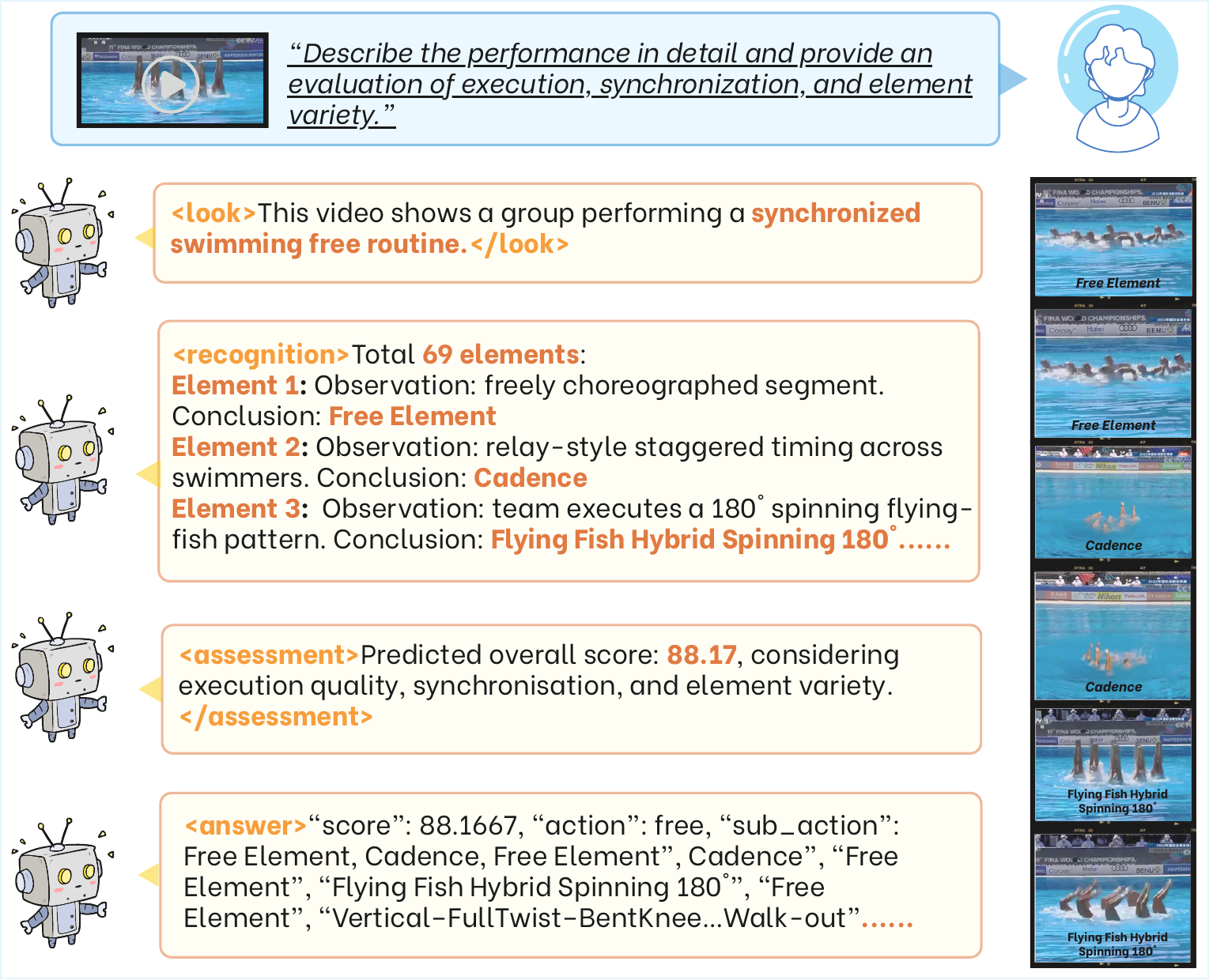}
  \caption{Step-by-step analysis of a synchronized swimming routine. The model takes video input and textual instruction, identifies fine-grained elements, evaluates performance quality across multiple dimensions, and outputs interpretable feedback.}
  \label{fig:vis}
\end{figure}

To address the limitations of direct score regression in interpretability and generalization, we propose \textbf{HieroAction} (see Fig.~\ref{fig:framework}), which formulates action quality assessment as a structured and hierarchical reasoning process. The model first extracts key motion cues from the input video, identifies the main actor and contextual information, and segments the continuous action into temporally coherent sub-actions, as shown in Fig.~\ref{fig:vis}. During inference, it follows a stepwise reasoning chain to analyze each sub-action, assess its quality using expert-defined criteria, and generate a structured output $\mathcal{A}$ including the action label, aligned score, and explanation. This hierarchical pipeline improves evaluation transparency and provides reliable feedback for downstream tasks.To train HieroAction, we build a hierarchical annotation set for each motion instance, including the overall label, sub-action segmentation, temporal boundaries, difficulty coefficient, and expert score. All question-answer pairs are organized by domain knowledge and follow the stepwise reasoning structure, reflecting the logic between action semantics and scoring rules. To improve language diversity and fluency, we use large language models (e.g., GPT) to paraphrase and polish the explanations without changing their meaning. The final dataset supports QA-style learning across various sports and motion understanding tasks. Training includes two stages: the first fine-tunes the Qwen-VL-2.5-7B model with supervised learning, and the second applies Hierarchical Policy Learning with multi-level rewards to enhance reasoning and generalization.

As discussed, \textbf{Stepwise Action Reasoning } and \textbf{Hierarchical Policy Learning} consist the key components of our method. The former helps generate diverse and interpretable training examples and guides the model to learn expert-aligned reasoning, and the later further optimizes the model's reasoning process and generalization ability via multi-level reward signals during reinforcement learning. In the following sections, we introduce the details of the two components.

\begin{tcolorbox}[
  title=Stepwise Action Reasoning Format,
  coltitle=white,
  colbacktitle=gray!70!black,
  colback=gray!5,
  colframe=black,
  boxrule=0.5pt,
  arc=3pt, outer arc=3pt,
  left=5pt, right=5pt, top=5pt, bottom=5pt,
  width=\linewidth,
  fonttitle=\bfseries\itshape,
  sharp corners,
  enhanced,
  breakable
]
\itshape
\textless look\textgreater\ Initial action contextualization \textless/look\textgreater\textbackslash n
\textless recognition\textgreater\ Phase: [functional role], Observation: [visual cue], Conclusion: [interpretation] \textless/recognition\textgreater\textbackslash n
\textless assessment\textgreater\ Performance evaluation and scoring justification \textless/assessment\textgreater\textbackslash n
\textless answer\textgreater\ Integrated reasoning synthesis and decision generation \textless/answer\textgreater
\end{tcolorbox}

\subsection{Stepwise Action Reasoning}
\label{sec:method_sar}

A key challenge in automated motion assessment is bridging the gap between low-level visual dynamics and the expert-level interpretability required for downstream applications such as skill evaluation and behavior analysis. To address this, we propose \textit{Stepwise Action Reasoning}, a principled reasoning paradigm that decomposes complex human actions into four sequential and interpretable stages. By explicitly modeling the expert evaluation process, SAR enables the model to produce fine-grained, transparent assessments that are both actionable and explainable. The structured format of this reasoning process is illustrated in the \textit{Stepwise Action Reasoning Format box}, which outlines the four stages of analysis.

\textit{Stage 1: Observation.}
The process begins with attentive observation, in which the model selectively focuses on the primary actor and relevant scene context, systematically filtering out background clutter and extraneous motion. This targeted perception ensures that subsequent inference is grounded in salient and meaningful visual evidence.

\textit{Stage 2: Recognition.}
The observed sequence is then temporally decomposed into semantically coherent sub-actions. For each segment, the model identifies its functional significance, distills key visual cues, and generates concise intermediate interpretations, thereby capturing the temporal and structural organization of the action.

\textit{Stage 3: Assessment.}
At this stage, the model evaluates the execution quality of each sub-action, considering factors such as difficulty, technical precision, and clarity of demonstration. Each local assessment is accompanied by a quantitative score and an explicit, human-interpretable rationale, closely emulating expert annotation practices.

\textit{Stage 4: Conclusion.}
Finally, SAR aggregates the stepwise assessments to form a holistic and consistent decision, producing a global judgment that is readily applicable to expert-oriented tasks while preserving the full reasoning trace for transparency and auditability.

To operationalize SAR in practice, we construct hierarchical, stepwise annotations that mirror this reasoning structure and leverage them to supervise model training. This annotation strategy not only grounds the model's predictions in expert-aligned logic but also facilitates effective learning of both the decomposition and inference processes intrinsic to SAR.  Fig. \ref{fig:framework} presents a representative example of the SAR process to facilitate intuitive understanding.

\subsection{Hierarchical Policy Learning}
\label{sec:method_rl}

To address the limitations of SFT, particularly its tendency to produce superficial pattern matching and limited generalization, we introduce hierarchical policy learning based on group relative policy optimization (GRPO). Supervised learning alone often fails to ensure that the model performs consistent, stepwise reasoning or aligns closely with expert evaluation standards. Our reinforcement learning framework is designed to explicitly encourage the model to generate structured, temporally precise, and semantically coherent assessments, as well as to produce scores that are well aligned with expert judgment. This is achieved through a set of targeted reward functions that promote structural fidelity, accurate temporal segmentation, robust semantic understanding, and precise scoring, thereby enhancing both the interpretability and reliability of the model's outputs. Specifically, there are 4 kinds of rewards.

\textbf{Format Reward.} To enforce structural consistency, we define the format reward $R_{\text{form}}$ to ensure the generated text includes all required reasoning stages—\texttt{<look>}, \texttt{<recognition>}, \texttt{<assessment>}, and \texttt{<answer>}—in correct sequence:
\[
R_{\text{form}} =
\begin{cases}
1, & \text{if all LRAA tags appear in correct order},\\
0, & \text{otherwise}.
\end{cases}
\]
This reward promotes adherence to our structured reasoning format, improving interpretability.

\textbf{Temporal Alignment Reward.} To ensure accurate temporal localization of sub-actions, we introduce the temporal alignment reward $R_{\text{temp}}$, calculated as the overlap between predicted and ground-truth segments:
\[
R_{\text{temp}} = \frac{1}{|\mathcal{M}|} \sum_{(i,j) \in \mathcal{M}} \text{IoU}(\tau_i^\star, \hat{\tau}_j)
\]
Here, $\tau_i^\star$ and $\hat{\tau}_j$ represent ground-truth and predicted intervals, respectively; $\mathcal{M}$ denotes optimal one-to-one segment matching, and $\text{IoU}(\cdot)$ indicates Intersection-over-Union. This reward emphasizes temporal precision during recognition.


\textbf{Hierarchical Action Reward.} To evaluate semantic accuracy at coarse and fine-grained levels, we define the hierarchical action reward $R_{\text{action}}$, balancing high-level action classification and low-level sub-action correctness:
\[
R_{\text{action}} = \alpha \cdot R_{\text{cls}} + (1 - \alpha) \cdot R_{\text{sub}}
\]
The parameter $\alpha \in [0,1]$ controls their relative importance; here we use $\alpha = 0.5$.

\indent$R_{\text{cls}}$ checks the predicted action label against the ground truth:
\[
R_{\text{cls}} = \mathbb{I}[\hat{y}_{\text{cls}} = y_{\text{cls}}]
\]
, where $\hat{y}_{\text{cls}}$ is the predicted action label and $y_{\text{cls}}$ is the ground truth.

$R_{\text{sub}}$ evaluates sub-action structure accuracy using normalized edit distance:
\[
R_{\text{sub}} = 1 - \frac{d_{\text{edit}}(\hat{S}, S)}{\max(|\hat{S}|, |S|)}
\]
, where $\hat{S}$ and $S$ are the predicted and ground-truth sub-action sequences, and $d_{\text{edit}}$ denotes the edit distance.

\textbf{Assessment Reward.} To align the model's outputs with expert scoring, we introduce the assessment reward $R_{\text{score}}$, measuring discrepancies between predicted and reference scores:
\[
R_{\text{score}} = \exp\left(-\lambda_{\text{score}} (q(y) - q^\star)^2 - \lambda_{\text{diff}} (d(y) - d^\star)^2 \right)
\]
, where $q(y)$ and $d(y)$ represent predicted quality and difficulty scores; $q^\star$ and $d^\star$ are ground-truth references. Parameters $\lambda_{\text{score}}$ and $\lambda_{\text{diff}}$ adjust sensitivity to these metrics, ensuring quantitative alignment with human assessments.

 In our implementation, all rewards are integrated into a single objective that jointly optimizes structure, timing, semantics, and scoring accuracy:
\[
\mathcal{R}(x,y) = \lambda_{\text{fmt}}R_{\text{form}} + \lambda_{\text{temp}}R_{\text{temp}} + \lambda_{\text{action}}R_{\text{action}} + \lambda_{\text{score}}R_{\text{score}}
\]
Each $\lambda$ parameter controls the weight of the respective reward. This combined formulation encourages consistent, precise, and interpretable reasoning across all stages. We use $0.1$ for the format reward and $0.3$ for others to maintain balance.

\begin{table*}[t]
\centering
\caption{Quantitative comparison with state-of-the-art methods on \textbf{FineDive}.}
\label{tab:sota_finedive}
\resizebox{0.75\textwidth}{!}{%
\begin{tabular}{c|cc|cc|cc|cc}
\toprule
\multirow{3}{*}{\centering\arraybackslash Methods} 
& \multicolumn{2}{c|}{Action Assessment} 
& \multicolumn{4}{c|}{Score Assessment} 
& \multicolumn{2}{c}{Content Quality} \\
\cmidrule{2-9}
& \multirow{2}{*}{\centering\arraybackslash Action Accuracy~$\uparrow$} 
& \multirow{2}{*}{\centering\arraybackslash SED~$\uparrow$} 
& \multicolumn{2}{c|}{Difficulty} 
& \multicolumn{2}{c|}{Score} 
& \multirow{2}{*}{\centering\arraybackslash Accuracy~$\uparrow$} 
& \multirow{2}{*}{\centering\arraybackslash Score~$\uparrow$} \\
& & & Spearman $\rho$~$\uparrow$ & R-l2~$\downarrow$ & Spearman $\rho$~$\uparrow$ & R-l2~$\downarrow$ & & \\
\midrule
\multicolumn{9}{c}{\textit{Action Quality Assessment Method}} \\
\midrule
USDL~\cite{tang2020uncertaintyawarescoredistributionlearning}   & -- & -- & -- & -- & 0.8302 & 0.0059 & -- & -- \\
CoRe~\cite{Yu_2021_ICCV}                                         & -- & -- & -- & -- & 0.8631 & 0.0056 & -- & -- \\
TSA~\cite{xu2022finedivingfinegraineddatasetprocedureaware}      & -- & -- & -- & -- & \textbf{0.8925} & \textbf{0.0048} & -- & -- \\
\midrule
\multicolumn{9}{c}{\textit{General Vision-Language Model}} \\
\midrule
NVILA~\cite{liu2025nvilaefficientfrontiervisual}               & 0     & 0     & 0.1061  & 0.3173  & -0.0165 & 0.7103  & 0.0158 & 1.40 \\
VideoLLaMA3~\cite{zhang2025videollama3frontiermultimodal}      & 0.0391 & 0.2979 & 0.0249  & 1.1712  & 0.0370  & 0.5856  & 0.0121 & 1.32 \\
InternVL-2.5~\cite{chen2024expanding}                          & 0.0733 & 0.3887 & 0.0697  & 0.2307  & -0.0341 & 0.2553  & 0.0384 & 1.32 \\
Gemini-2.5-Flash~\cite{gemini2_5_flash}                      & 0.0041 & 0.3702 & 0.1310  & 0.2531  & 0.0646  & 0.2620  & 0.0186 & 1.34 \\
Qwen-VL-2.5~\cite{qwen2.5-VL}                                                    & 0.2564 & 0.2823 & 0.0263  & 0.3885  & 0.0400  & 0.2787  & 0.0290 & 1.39 \\
VideoLLaMA3-SFT                                              & 0.8987     & 0.9468     & 0.9078      & 0.0542      & 0.7875      & 0.1253      & 0.8985     & 3.81  \\
InternVL-2.5-SFT                                             & 0.9023     & 0.9483     & 0.9161      & 0.0452      & 0.8148      & 0.0886      & 0.9012     & 3.84  \\
\midrule
\multicolumn{9}{c}{\textit{RL-based Method}} \\
\midrule
VideoR1~\cite{feng2025videor1reinforcingvideoreasoning}        & 0      & 0.1015 & -0.0483 & 0.2508  & -0.0657 & 0.1722  & 0.0187     & 1.28  \\
VideoChat-R1~\cite{li2025videochatr1enhancingspatiotemporalperception} & 0      & 0.1415 & -0.0232 & 0.2285  & -0.0308 & 0.1896  & 0.0267     & 1.35  \\
VideoChat-R1-SFT                                              & 0.8655     & 0.9412     & 0.9071      & 0.0664      & 0.8363      & 0.0801      & 0.8945     & 3.72  \\
\midrule
\textbf{Ours} & \textbf{0.9344} & \textbf{0.9731} & \textbf{0.9423} & \textbf{0.0158} & 0.8564 & 0.0714 & \textbf{0.9279} & \textbf{4.12} \\
\bottomrule
\end{tabular}
}
\end{table*}

\section{Experiment}

\subsection{Implementation Details}

\textbf{Training settings.} We use Qwen2.5-VL-Instruct-7B~\cite{qwen2.5-VL} as the base model, trained with SAR-formatted data (\textless look\textgreater, \textless recognition\textgreater, \textless assessment\textgreater, \textless answer\textgreater). Supervised fine-tuning is performed on 4$\times$A100 (40GB) GPUs, batch size 16, learning rate $2 \times 10^{-5}$, for 25 epochs, with a maximum output length of 3072 tokens and greedy decoding. For SAR, we design SAR-based rewards for ranking consistency, score accuracy, and explanation quality. For each input, $G=8$ responses are sampled, and the highest-reward response is used for training. HPL follows the same setup as SFT, with a KL coefficient $\beta=0.04$, but uses a smaller learning rate $1 \times 10^{-6}$.

\noindent \textbf{Datasets.} We evaluate on three fine-grained action analysis benchmarks: FineDive~\cite{xu2022finedivingfinegraineddatasetprocedureaware}, FineFS~\cite{JI2023FineFS}, and LOGO~\cite{zhang2024logolongformvideodataset}. FineDive contains 3,000 diving videos (10–20s each), split into 2,251 for training and 749 for testing; it features high temporal density and subtle class differences. FineFS includes 1,167 figure skating videos (2–4 min each), with 933 for training and 234 for testing, challenging models to localize sparse sub-actions in long sequences. LOGO comprises 200 artistic swimming videos (150 train, 50 test), focusing on multi-person coordination and sparse group actions. These datasets span short, dense individual actions (FineDive), long, sparse performances (FineFS), and complex group dynamics (LOGO), enabling comprehensive evaluation across temporal and interaction dimensions.

\subsection{Metrics for Fine-Grained Action Analysis}
\label{sec:metrics}

\subsection{Evaluation Protocol}

To comprehensively assess model performance, we employ a unified evaluation protocol spanning three aspects: \textbf{motion fidelity}, \textbf{scoring precision}, and \textbf{semantic correctness}.

\textbf{Motion Fidelity.}  
We evaluate the model’s ability to recognize action categories and capture sub-action structures. 
Action accuracy ($\text{Acc}_{\text{action}}$) measures correct classification:
\begin{equation}
\text{Acc}_{\text{action}} = \frac{1}{N} \sum_{i=1}^{N} \mathbb{I}\left[\hat{y}_i = y_i^\star\right]
\end{equation}
Following~\cite{ding2023temporalactionsegmentationanalysis}, Sub-action Edit Distance (SED) quantifies sequence similarity by computing the normalized edit distance between predicted and ground-truth sub-action sequences.
\begin{equation}
\text{SED} = 1 - \frac{\text{EditDistance}(G, P)}{\max\left(|G|, |P|\right)}
\end{equation}
, where $G$ and $P$ are ground-truth and predicted sub-action sequences.

\textbf{Scoring Precision.}  
This aspect covers both ranking consistency and absolute score accuracy. Following~\cite{9009513,xu2022finedivingfinegraineddatasetprocedureaware,parmar2019performedmultitasklearningapproach}, we adopt Spearman’s rank correlation coefficient ($\rho$) to assess the degree of monotonic agreement between predicted and ground-truth rankings:
\begin{equation}
\rho = \frac{\sum_{i=1}^{N} (y_i - \bar{y})(\hat{y}_i - \bar{\hat{y}})}{\sqrt{\sum_{i=1}^{N} (y_i - \bar{y})^2} \sqrt{\sum_{i=1}^{N} (\hat{y}_i - \bar{\hat{y}})^2}}
\end{equation}
, where $y_i$ and $\hat{y}_i$ denote the ground-truth and predicted ranks for the $i$-th sample, and $\bar{y}$ and $\bar{\hat{y}}$ are their respective means. Higher $\rho$ indicates better ranking performance.

We also follow~\cite{yu2021groupawarecontrastiveregressionaction} to adopt relative $\ell_2$-distance (R-$\ell_2$) for normalized score error evaluation. Given the maximum and minimum ground-truth scores of an action, denoted by $y_{\max}$ and $y_{\min}$, R-$\ell_2$ is defined as:
\begin{equation}
\text{R-}\ell_2 = \frac{1}{N} \sum_{i=1}^{N} \left( \frac{|y_i - \hat{y}_i|}{y_{\max} - y_{\min}} \right)
\end{equation}
, where $y_i$ and $\hat{y}_i$ represent the ground-truth and predicted scores of the $i$-th sample. A lower R-$\ell_2$ indicates better score accuracy.

\textbf{Semantic Correctness.}  
Following~\cite{chen2024motionllm}, we evaluate the alignment with expert reasoning using an LLM-based judge by assessing (i) the motion similarity reflected in sub-action descriptions, and (ii) the overall correctness of the decision, encompassing the predicted action label, score, and accompanying justification.

\begin{table}[htbp]
\centering
\caption{Quantitative comparison with state-of-the-art methods on \textbf{FineFs}.}
\label{tab:finefs_singlecol}
\resizebox{\linewidth}{!}{%
\begin{tabular}{l|c|cc|cc}
\toprule
\multirow{2}{*}{\textbf{Methods}} 
& \textbf{Action Assessment} 
& \multicolumn{2}{c|}{\textbf{Score Assessment}} 
& \multicolumn{2}{c}{\textbf{Content Quality}} \\
\cmidrule{2-6}
& \textbf{SED~$\uparrow$} 
& \textbf{Spearman $\rho$~$\uparrow$} & \textbf{R-l2~$\downarrow$} 
& \textbf{Acc~$\uparrow$} & \textbf{Score~$\uparrow$} \\
\midrule
\multicolumn{6}{c}{\textit{Action Quality Assessment Methods}} \\
\midrule
MS-LSTM~\cite{ma2024mslstmexploringspatiotemporalmultiscale}     & --      & 0.5795      & --     & --     & -- \\
TSA~\cite{xu2022finedivingfinegraineddatasetprocedureaware}      & --      & 0.7005      & --     & --     & -- \\
GDLT~\cite{gdlt22}                                                & --      & \textbf{0.7240}      & --     & --     & -- \\
\midrule
\multicolumn{6}{c}{\textit{General Vision-Language Model}} \\
\midrule
NVILA~\cite{liu2025nvilaefficientfrontiervisual}               & 0.0007  & 0.0908  & 0.6474  & 0.0000 & 1.10 \\
VideoLLaMA3~\cite{zhang2025videollama3frontiermultimodal}      & 0.0625  & 0.0953  & 0.3889  & 0.0000 & 1.13 \\
InternVL-2.5~\cite{chen2024expanding}                           & 0.0055  & -0.0864 & 0.6434  & 0.0218 & 1.63 \\
Gemini-2.5-Flash~\cite{gemini2_5_flash}                       & 0.5515  & 0.4440  & 0.3529  & 0.1009 & 2.11 \\
Qwen-VL-2.5~\cite{qwen2.5-VL}                                                      & 0.0073  & -0.0982 & 0.5921  & 0.0133 & 1.32 \\
VideoLLaMA3-SFT                                                & 0.6421      & 0.2653      & 0.3732      & 0.4615     & 2.85 \\
InternVL-2.5-SFT                                               & 0.6264      & 0.2581      & 0.3884      & 0.4487     & 2.83 \\
\midrule
\multicolumn{6}{c}{\textit{RL-based Method}} \\
\midrule
VideoR1~\cite{feng2025videor1reinforcingvideoreasoning}         & 0.0268  & -0.0892 & 0.3064  & 0.0000     & 1.23 \\
VideoChat-R1~\cite{li2025videochatr1enhancingspatiotemporalperception} & 0.0580  & -0.1841 & 0.2991  & 0.0171     & 1.37 \\
VideoChat-R1-SFT                                                & 0.6123      & 0.2483      & 0.3964      & 0.4273     & 2.72 \\
\midrule
\textbf{Ours}                                                   & \textbf{0.6761}  & 0.4232  & \textbf{0.1136}  & \textbf{0.5556} & \textbf{3.48} \\
\bottomrule
\end{tabular}
}
\end{table}

\begin{table}[htbp]
\centering
\caption{Quantitative comparison with state-of-the-art methods on \textbf{LOGO}.}
\label{tab:logo_singlecol}
\resizebox{\linewidth}{!}{%
\begin{tabular}{l|c|cc|cc}
\toprule
\multirow{2}{*}{\textbf{Methods}} 
& \textbf{Action Assessment} 
& \multicolumn{2}{c|}{\textbf{Score Assessment}} 
& \multicolumn{2}{c}{\textbf{Content Quality}} \\
\cmidrule{2-6}
& \textbf{SED~$\uparrow$} 
& \textbf{Spearman $\rho$~$\uparrow$} & \textbf{R-l2~$\downarrow$} 
& \textbf{Acc~$\uparrow$} & \textbf{Score~$\uparrow$} \\
\midrule
\multicolumn{6}{c}{\textit{Action Quality Assessment Methods}} \\
\midrule
USDL~\cite{tang2020uncertaintyawarescoredistributionlearning}   & --      & 0.4725  & \textbf{0.0508} & --     & -- \\
CoRe~\cite{Yu_2021_ICCV}                                        & --      & \textbf{0.5002}  & 0.0596 & --     & -- \\
TSA~\cite{xu2022finedivingfinegraineddatasetprocedureaware}     & --      & 0.4751  & 0.4778 & --     & -- \\
\midrule
\multicolumn{6}{c}{\textit{General Vision-Language Model}} \\
\midrule
NVILA~\cite{liu2025nvilaefficientfrontiervisual}               & 0.0110  & N/A     & 0.2699 & 0.0000 & 1.58 \\
VideoLLaMA3~\cite{zhang2025videollama3frontiermultimodal}      & 0.0158  & -0.1392 & 0.7546 & 0.0000 & 1.36 \\
InternVL-2.5~\cite{chen2024expanding}                          & 0.0166  & N/A     & 0.2691 & 0.0000 & 1.70 \\
Gemini-2.5-Flash~\cite{gemini2_5_flash}                      & 0.0023  & 0.2898  & 0.2746 & 0.1837 & 2.33 \\
Qwen-VL-2.5~\cite{qwen2.5-VL}                                                     & 0.0158  & 0.1684  & 0.2417 & 0.0200 & 1.78 \\
VideoLLaMA3-SFT                                               & 0.4954      & 0.0831      & 0.3694     & 0.2200     & 2.12 \\
InternVL-2.5-SFT                                              & 0.5071      & 0.1038      & 0.3507     & 0.2400     & 2.17 \\
\midrule
\multicolumn{6}{c}{\textit{RL-based Method}} \\
\midrule
VideoR1~\cite{feng2025videor1reinforcingvideoreasoning}        & 0.0213  & -0.0940 & 0.2235 & 0.0000     & 1.54 \\
VideoChat-R1~\cite{li2025videochatr1enhancingspatiotemporalperception} & 0.0166  & N/A     & 0.2433 & 0.0000     & 1.43 \\
VideoChat-R1-SFT                                               & 0.4931  & 0.0211  & 0.3805 & 0.2200     & 2.08 \\
\midrule
\textbf{Ours}           & \textbf{0.5793}     & 0.3441     & 0.2103                                       & \textbf{0.4200}      & \textbf{3.03}       \\
\bottomrule
\end{tabular}
}
\end{table}

\begin{table*}[t]
\centering
\caption{Quantitative results of the ablation study}
\label{tab:ablation_study_all}
\resizebox{\textwidth}{!}{%
\begin{tabular}{c|cccc|cc|cc|cc|c|c}
\toprule
\multirow{3}{*}{\textbf{Dataset}} 
& \multicolumn{4}{c|}{\textbf{Model Variant}} 
& \multicolumn{2}{c|}{\textbf{Action Assessment}} 
& \multicolumn{4}{c|}{\textbf{Score Assessment}} 
& \multicolumn{2}{c}{\textbf{Content Quality}} \\
\cmidrule{2-13}
& \multirow{2}{*}{\textbf{w/o CoT}} 
& \multirow{2}{*}{\textbf{CoT}} 
& \multirow{2}{*}{\textbf{SAR}} 
& \multirow{2}{*}{\textbf{RL}} 
& \multirow{2}{*}{\textbf{Action Accuracy~$\uparrow$}} 
& \multirow{2}{*}{\textbf{SED~$\uparrow$}} 
& \multicolumn{2}{c|}{\textbf{Difficulty}} 
& \multicolumn{2}{c|}{\textbf{Score}} 
& \multirow{2}{*}{\textbf{Accuracy~$\uparrow$}} 
& \multirow{2}{*}{\textbf{Score~$\uparrow$}} \\
\cmidrule{8-11}
& & & & & & & \textbf{Spearman $\rho$~$\uparrow$} & \textbf{R-l2~$\downarrow$} & \textbf{Spearman $\rho$~$\uparrow$} & \textbf{R-l2~$\downarrow$} & & \\
\midrule

\multirow{4}{*}{FineDive}
& \checkmark &             &             &             & 0.8798 & 0.9477 & 0.8953 & 0.0818 & 0.8153 & 0.0856 & 0.8171   & 3.31 \\
&            & \checkmark  &             &             & 0.8945 & 0.9551 & 0.9297 & 0.0471  & 0.8371 & 0.0825  & 0.8678   & 3.76 \\
&            &             & \checkmark  &             & 0.9052 & 0.9564 & 0.9138 & 0.0383  & 0.8011 & 0.0975 & 0.9080   & 3.90 \\
&            &             & \checkmark  & \checkmark  & \textbf{0.9344} & \textbf{0.9731} & \textbf{0.9423} & \textbf{0.0158} & \textbf{0.8564} & \textbf{0.0714} & \textbf{0.9279} & \textbf{4.12} \\
\midrule

\multirow{4}{*}{FineFs}
& \checkmark &             &             &             & --     & 0.6167     & --     & --     & 0.3315     & 0.2245     & 0.2234   & 1.89   \\
&            & \checkmark  &             &             & --     & 0.6203     & --     & --     & 0.2725     & 0.3476     & 0.4276   & 2.67   \\
&            &             & \checkmark  &             & --     & 0.6326     & --     & --     & 0.2553     & 0.3778     & 0.4530   & 2.81   \\
&            &             & \checkmark  & \checkmark  & --     & \textbf{0.6761} & --     & --     & \textbf{0.4232} & \textbf{0.1136} & \textbf{0.5556} & \textbf{3.48}   \\
\midrule

\multirow{4}{*}{LOGO}
& \checkmark &             &             &             & --     & 0.5146 & --     & --     & 0.2551  & 0.3212  & 0.1600   & 1.71 \\
&            & \checkmark  &             &             & --     & 0.5024 & --     & --     & 0.0525  & 0.3720  & 0.2200   & 1.91 \\
&            &             & \checkmark  &             & --     & 0.4993 & --     & --     & -0.0057 & 0.3321  & 0.2600   & 2.23 \\
&            &             & \checkmark  & \checkmark  & --     & \textbf{0.5703} & --     & --     & \textbf{0.3441} & \textbf{0.2103} & \textbf{0.4200} & \textbf{3.03} \\

\bottomrule
\end{tabular}
}
\end{table*}

\subsection{Comparison with SOTA Methods}

We compare our method with a range of recent and representative baselines. These include state-of-the-art general-purpose vision-language (VL) models—Gemini-2.5-Flash~\cite{gemini2_5_flash}, InternVL-2.5~\cite{chen2024expanding}, NVILA~\cite{liu2025nvilaefficientfrontiervisual}, and VideoLLaMA-3~\cite{zhang2025videollama3frontiermultimodal}—as well as leading reinforcement learning-based VL models such as VideoR1~\cite{feng2025videor1reinforcingvideoreasoning} and VideoChat-R1~\cite{li2025videochatr1enhancingspatiotemporalperception}. To further validate the effectiveness of our fine-grained evaluation metrics, we also include specialized Action Quality Assessment (AQA) models focused on scoring performance, namely USDL~\cite{tang2020uncertaintyawarescoredistributionlearning}, CoRe~\cite{Yu_2021_ICCV}, TSA~\cite{xu2022finedivingfinegraineddatasetprocedureaware}, MS-LSTM~\cite{ma2024mslstmexploringspatiotemporalmultiscale}, and GDLT~\cite{gdlt22}. The results are as follows.

\subsubsection{Zero-Shot Results.}

We evaluate the zero-shot performance of pre-trained models on our benchmarks, as shown in Tables~\ref{tab:sota_finedive}, \ref{tab:finefs_singlecol}, and \ref{tab:logo_singlecol}. General vision-language models can recognize coarse actions (e.g., diving, figure skating) but struggle with fine-grained categories and official codes, showing limited domain knowledge. They also lack reliable sub-action modeling, especially on longer videos, where outputs often lack structure or contain formatting errors. For scoring, most baselines produce inconsistent and non-informative results. While models like Gemini perform better on FineFS, overall prediction remains weak. RL-based models, though designed for sequential reasoning, generalize poorly due to task-specific objectives and reward dependence, and often underperform general models in zero-shot settings.

\subsubsection{Supervised Adaptation.}

We further conduct supervised fine-tuning on vision-language models. The corresponding results are presented in the SFT rows of the General Vision-Language Models and RL-based Methods sections in Tables~\ref{tab:sota_finedive}, \ref{tab:finefs_singlecol}, and \ref{tab:logo_singlecol}. Fine-tuned general VL models show significant gains, generating more structured and coherent predictions. Models can decompose complex actions, capture temporal dependencies, and associate motion cues with scoring criteria, resulting in improved accuracy and interpretability. However, they still lag behind specialized Action Quality Assessment (AQA) models on fine-grained metrics. RL-based VL models, after fine-tuning, also improve but remain less competitive than fine-tuned general VL models, likely due to pre-training on narrow, task-specific objectives.

\subsubsection{Comparison with AQA Methods.}
We also compare with state-of-the-art AQA models that output a single motion quality score, as shown in Tables~\ref{tab:sota_finedive}, \ref{tab:finefs_singlecol}, and \ref{tab:logo_singlecol}. These methods perform well on fine-grained metrics (e.g., Spearman correlation) but lack interpretability—they do not reveal decision logic or sub-action recognition, limiting their use in tasks like coaching. In contrast, our model achieves comparable fine-grained performance (e.g., on FineDive) while providing structured and interpretable evaluations with sub-action decomposition, which are essential for real-world applications. Still, further refinement is needed to fully outperform specialized AQA methods.

In summary, HieroAction advances fine-grained sports analysis by integrating stepwise action reasoning and hierarchical policy learning. It enables the model to interpret actions progressively and capture critical temporal cues. Compared to existing baselines, our approach consistently achieves higher fine-grained accuracy and interpretability, underscoring the benefits of structured reasoning and policy learning.

\subsection{Ablation Study}
This section primarily investigates the impact of our two key components: Stepwise Action Reasoning and Hierarchical Policy Learning.

\paragraph{Ablation on Stepwise Action Reasoning.} To assess the effect of SAR, we evaluate three variants: (1)~No CoT, where the model predicts scores directly without reasoning; (2)~Standard CoT (Think-then-Answer), which uses a single-step format that often mixes different types of information; and (3)~Stepwise Action Reasoning, which adopts a structured, multi-step format with separated stages. As shown in Table~\ref{tab:ablation_study_all}, SAR achieves better fine-grained accuracy and interpretability on FineDive by breaking down complex actions into sub-stages. On long-form datasets like FineFS and LOGO, both stepwise and standard CoT improve interpretability. However, as action complexity increases, longer reasoning sequences may impair supervised training, and direct answers sometimes yield better fine-grained results. This suggests that supervised fine-tuning alone is insufficient for handling complex, extended sequences.

\paragraph{Ablation on Hierarchical Policy Learning.} We further ablate the proposed {Hierarchical Policy Learning} strategy to assess its effectiveness. This approach introduces multiple reward functions, each supervising a specific component of the Stepwise Action Reasoning process, including temporal modeling, sub-action recognition, and score prediction. These rewards operate in a coordinated manner, providing fine-grained supervision across different reasoning stages. Experimental results in Table \ref{tab:ablation_study_all} demonstrate that incorporating these specialized rewards significantly improves fine-grained metrics over the baseline, confirming the effectiveness of Hierarchical Policy Learning for complex action understanding tasks.

\section{Conclusion}
We propose HieroAction, a vision-language model that combines stepwise action reasoning with hierarchical policy learning for human action assessment. Our method extracts key motion features, uses temporal reasoning, and follows expert-guided prompts to give step-by-step and interpretable evaluations. The hierarchical design helps the model focus on fine-grained details, break down complex actions into clear phases, and find where mistakes happen. As a result, the evaluations are more accurate, transparent, and reliable. Experiments on several datasets show strong performance across different metrics. This work shows how combining reasoning and learning in a vision-language framework can improve understanding of human actions.

\bibliography{aaai2026}

\begin{thebibliography}{39}
\providecommand{\natexlab}[1]{#1}

\bibitem[{Bai et~al.(2025)Bai, Li, Liu, Tang, Zhang, Sun, Chu, and Tang}]{bai2025univgr1reasoningguideduniversal}
Bai, S.; Li, M.; Liu, Y.; Tang, J.; Zhang, H.; Sun, L.; Chu, X.; and Tang, Y. 2025.
\newblock UniVG-R1: Reasoning Guided Universal Visual Grounding with Reinforcement Learning.
\newblock arXiv:2505.14231.

\bibitem[{Bianchi and Liotta(2025)}]{bianchi2025patsproficiencyawaretemporalsampling}
Bianchi, E.; and Liotta, A. 2025.
\newblock PATS: Proficiency-Aware Temporal Sampling for Multi-View Sports Skill Assessment.
\newblock arXiv:2506.04996.

\bibitem[{Chen et~al.(2024{\natexlab{a}})Chen, Lu, Zeng, Zhang, Wang, Zhang, and Zhang}]{chen2024motionllm}
Chen, L.-H.; Lu, S.; Zeng, A.; Zhang, H.; Wang, B.; Zhang, R.; and Zhang, L. 2024{\natexlab{a}}.
\newblock MotionLLM: Understanding Human Behaviors from Human Motions and Videos.
\newblock arXiv:2405.20340.

\bibitem[{Chen et~al.(2025)Chen, Wang, Cao, Liu, Gao, Cui, Zhu, Ye, Tian, Liu, Gu, Wang, Li, Ren, Chen, Luo, Wang, Jiang, Wang, He, Shi, Zhang, Lv, Wang, Shao, Chu, Tu, He, Wu, Deng, Ge, Chen, Zhang, Wang, Dou, Lu, Zhu, Lu, Lin, Qiao, Dai, and Wang}]{chen2024expanding}
Chen, Z.; Wang, W.; Cao, Y.; Liu, Y.; Gao, Z.; Cui, E.; Zhu, J.; Ye, S.; Tian, H.; Liu, Z.; Gu, L.; Wang, X.; Li, Q.; Ren, Y.; Chen, Z.; Luo, J.; Wang, J.; Jiang, T.; Wang, B.; He, C.; Shi, B.; Zhang, X.; Lv, H.; Wang, Y.; Shao, W.; Chu, P.; Tu, Z.; He, T.; Wu, Z.; Deng, H.; Ge, J.; Chen, K.; Zhang, K.; Wang, L.; Dou, M.; Lu, L.; Zhu, X.; Lu, T.; Lin, D.; Qiao, Y.; Dai, J.; and Wang, W. 2025.
\newblock Expanding Performance Boundaries of Open-Source Multimodal Models with Model, Data, and Test-Time Scaling.
\newblock arXiv:2412.05271.

\bibitem[{Chen et~al.(2024{\natexlab{b}})Chen, Zhou, Shen, Hong, Sun, Gutfreund, and Gan}]{chen2024vcot}
Chen, Z.; Zhou, Q.; Shen, Y.; Hong, Y.; Sun, Z.; Gutfreund, D.; and Gan, C. 2024{\natexlab{b}}.
\newblock Visual Chain-of-Thought Prompting for Knowledge-Based Visual Reasoning.
\newblock \emph{AAAI}, 1254--1262.

\bibitem[{Dave, Rizve, and Shah(2024)}]{dave2025finepseudo}
Dave, I.~R.; Rizve, M.~N.; and Shah, M. 2024.
\newblock FinePseudo: Improving Pseudo-labelling Through Temporal-Alignablity for Semi-supervised Fine-Grained Action Recognition.
\newblock In \emph{ECCV}, 389--408.

\bibitem[{DeepMind(2025)}]{gemini2_5_flash}
DeepMind, G. 2025.
\newblock Gemini 2.5 Flash.

\bibitem[{Ding, Sener, and Yao(2024)}]{ding2023temporalactionsegmentationanalysis}
Ding, G.; Sener, F.; and Yao, A. 2024.
\newblock Temporal Action Segmentation: An Analysis of Modern Techniques.
\newblock \emph{IEEE Trans. Pattern Anal. Mach. Intell.}, 1011--1030.

\bibitem[{Feng et~al.(2025)Feng, Gong, Li, Guo, Wang, Peng, Wu, Zhang, Wang, and Yue}]{feng2025videor1reinforcingvideoreasoning}
Feng, K.; Gong, K.; Li, B.; Guo, Z.; Wang, Y.; Peng, T.; Wu, J.; Zhang, X.; Wang, B.; and Yue, X. 2025.
\newblock Video-R1: Reinforcing Video Reasoning in MLLMs.
\newblock arXiv:2503.21776.

\bibitem[{Han et~al.(2025)Han, Zhou, Atapour-Abarghouei, Liang, and Shum}]{han2025finecausalcausalbasedframeworkinterpretable}
Han, R.; Zhou, K.; Atapour-Abarghouei, A.; Liang, X.; and Shum, H. P.~H. 2025.
\newblock FineCausal: A Causal-Based Framework for Interpretable Fine-Grained Action Quality Assessment.
\newblock In \emph{CVPR Workshops}, 6018--6027.

\bibitem[{Ji et~al.(2023)Ji, Ye, Huang, Mao, Zhou, and Gao}]{JI2023FineFS}
Ji, Y.; Ye, L.; Huang, H.; Mao, L.; Zhou, Y.; and Gao, L. 2023.
\newblock Localization-assisted Uncertainty Score Disentanglement Network for Action Quality Assessment.
\newblock In \emph{ACM MM}, 8590–8597.

\bibitem[{Jiang et~al.(2025)Jiang, Wang, Salekin, Atighehchian, and Zhang}]{jiang2025domainadaptationvlmsoccer}
Jiang, T.; Wang, H.; Salekin, M.~S.; Atighehchian, P.; and Zhang, S. 2025.
\newblock Domain Adaptation of VLM for Soccer Video Understanding.
\newblock In \emph{CVPR Workshops}, 6111--6121.

\bibitem[{Kojima et~al.(2022)Kojima, Gu, Reid, Matsuo, and Iwasawa}]{kojima2022zero}
Kojima, T.; Gu, S.~S.; Reid, M.; Matsuo, Y.; and Iwasawa, Y. 2022.
\newblock Large Language Models are Zero-Shot Reasoners.
\newblock In \emph{NeurIPS}, 22199--22213.

\bibitem[{Li et~al.(2025{\natexlab{a}})Li, Yan, Meng, Dong, Zeng, He, Wang, Qiao, Wang, and Wang}]{li2025videochatr1enhancingspatiotemporalperception}
Li, X.; Yan, Z.; Meng, D.; Dong, L.; Zeng, X.; He, Y.; Wang, Y.; Qiao, Y.; Wang, Y.; and Wang, L. 2025{\natexlab{a}}.
\newblock VideoChat-R1: Enhancing Spatio-Temporal Perception via Reinforcement Fine-Tuning.
\newblock arXiv:2504.06958.

\bibitem[{Li et~al.(2025{\natexlab{b}})Li, Chen, Li, Liu, Wang, Luo, Hu, and Zhang}]{li2025veripocultivatinglongreasoning}
Li, Y.; Chen, X.; Li, Z.; Liu, Z.; Wang, L.; Luo, W.; Hu, B.; and Zhang, M. 2025{\natexlab{b}}.
\newblock VerIPO: Cultivating Long Reasoning in Video-LLMs via Verifier-Gudied Iterative Policy Optimization.
\newblock arXiv:2505.19000.

\bibitem[{Liu et~al.(2025)Liu, Zhu, Shi, Zhang, Lou, Yang, Xi, Cao, Gu, Li, Li, Tang, Fang, Chen, Hsieh, Huang, Cheng, Hu, Liu, Krishna, Molchanov, Kautz, Yin, Han, and Lu}]{liu2025nvilaefficientfrontiervisual}
Liu, Z.; Zhu, L.; Shi, B.; Zhang, Z.; Lou, Y.; Yang, S.; Xi, H.; Cao, S.; Gu, Y.; Li, D.; Li, X.; Tang, H.; Fang, Y.; Chen, Y.; Hsieh, C.-Y.; Huang, D.-A.; Cheng, A.-C.; Hu, J.; Liu, S.; Krishna, R.; Molchanov, P.; Kautz, J.; Yin, H.; Han, S.; and Lu, Y. 2025.
\newblock NVILA: Efficient Frontier Visual Language Models.
\newblock In \emph{CVPR}, 4122--4134.

\bibitem[{Ma, Zhang, and Liu(2023)}]{ma2024mslstmexploringspatiotemporalmultiscale}
Ma, Z.; Zhang, H.; and Liu, J. 2023.
\newblock MS-LSTM: Exploring spatiotemporal multiscale representations in video prediction domain.
\newblock \emph{Appl. Soft Comput.}

\bibitem[{Pan, Gao, and Zheng(2019)}]{9009513}
Pan, J.-H.; Gao, J.; and Zheng, W.-S. 2019.
\newblock Action Assessment by Joint Relation Graphs.
\newblock In \emph{ICCV}, 6330--6339.

\bibitem[{Park et~al.(2025)Park, Kim, Kim, Kim, and Ro}]{park2025dipr1deepinspectionperception}
Park, S.; Kim, H.; Kim, J.; Kim, S.; and Ro, Y.~M. 2025.
\newblock DIP-R1: Deep Inspection and Perception with RL Looking Through and Understanding Complex Scenes.
\newblock arXiv:2505.23179.

\bibitem[{Parmar and Morris(2019)}]{parmar2019performedmultitasklearningapproach}
Parmar, P.; and Morris, B.~T. 2019.
\newblock What and How Well You Performed? A Multitask Learning Approach to Action Quality Assessment.
\newblock In \emph{CVPR}.

\bibitem[{Pirsiavash, Vondrick, and Torralba(2014)}]{pirsiavash2014assessing}
Pirsiavash, H.; Vondrick, C.; and Torralba, A. 2014.
\newblock Assessing the Quality of Actions.
\newblock In \emph{ECCV}, 556--571.

\bibitem[{Reilly et~al.(2025)Reilly, Chakraborty, Sinha, Govind, Wang, Bremond, Xue, and Das}]{reilly2025llavidallargelanguagevision}
Reilly, D.; Chakraborty, R.; Sinha, A.; Govind, M.~K.; Wang, P.; Bremond, F.; Xue, L.; and Das, S. 2025.
\newblock LLAVIDAL: A Large LAnguage VIsion Model for Daily Activities of Living.
\newblock In \emph{CVPR}, 24297--24308.

\bibitem[{Rudin(2019)}]{rudin2019interpretable}
Rudin, C. 2019.
\newblock Stop Explaining Black Box Machine Learning Models for High Stakes Decisions and Use Interpretable Models Instead.
\newblock \emph{Nat. Mach. Intell.}, 206--215.

\bibitem[{Shao et~al.(2024)Shao, Qian, Xiao, Song, Zong, Wang, Liu, and Li}]{shao2024visualcotadvancingmultimodal}
Shao, H.; Qian, S.; Xiao, H.; Song, G.; Zong, Z.; Wang, L.; Liu, Y.; and Li, H. 2024.
\newblock Visual CoT: Advancing Multi-Modal Language Models with a Comprehensive Dataset and Benchmark for Chain-of-Thought Reasoning.
\newblock In \emph{NeurIPS}, 8612--8642.

\bibitem[{Tang et~al.(2020)Tang, Ni, Zhou, Zhang, Lu, Wu, and Zhou}]{tang2020uncertaintyawarescoredistributionlearning}
Tang, Y.; Ni, Z.; Zhou, J.; Zhang, D.; Lu, J.; Wu, Y.; and Zhou, J. 2020.
\newblock Uncertainty-Aware Score Distribution Learning for Action Quality Assessment.
\newblock In \emph{CVPR}.

\bibitem[{Team(2025)}]{qwen2.5-VL}
Team, Q. 2025.
\newblock Qwen2.5-VL.

\bibitem[{Wei et~al.(2022)Wei, Wang, Schuurmans, Bosma, Ichter, Xia, Chi, Le, and Zhou}]{wei2022chain}
Wei, J.; Wang, X.; Schuurmans, D.; Bosma, M.; Ichter, B.; Xia, F.; Chi, E.; Le, Q.~V.; and Zhou, D. 2022.
\newblock Chain-of-Thought Prompting Elicits Reasoning in Large Language Models.
\newblock In \emph{NeurIPS}, 24824--24837.

\bibitem[{Xu, Zeng, and Zheng(2022)}]{gdlt22}
Xu, A.; Zeng, L.-A.; and Zheng, W.-S. 2022.
\newblock Likert Scoring with Grade Decoupling for Long-term Action Assessment.
\newblock In \emph{CVPR}, 3222--3231.

\bibitem[{Xu et~al.(2024)Xu, Ke, Li, Xu, Wu, Lin, and Guo}]{xu2025visionlanguageaqa}
Xu, H.; Ke, X.; Li, Y.; Xu, R.; Wu, H.; Lin, X.; and Guo, W. 2024.
\newblock Vision-Language Action Knowledge Learning for Semantic-Aware Action Quality Assessment.
\newblock In \emph{ECCV}, 423--440.

\bibitem[{Xu et~al.(2022)Xu, Rao, Yu, Chen, Zhou, and Lu}]{xu2022finedivingfinegraineddatasetprocedureaware}
Xu, J.; Rao, Y.; Yu, X.; Chen, G.; Zhou, J.; and Lu, J. 2022.
\newblock FineDiving: A Fine-Grained Dataset for Procedure-Aware Action Quality Assessment.
\newblock In \emph{CVPR}, 2949--2958.

\bibitem[{Yu et~al.(2021{\natexlab{a}})Yu, Rao, Zhao, Lu, and Zhou}]{Yu_2021_ICCV}
Yu, X.; Rao, Y.; Zhao, W.; Lu, J.; and Zhou, J. 2021{\natexlab{a}}.
\newblock Group-Aware Contrastive Regression for Action Quality Assessment.
\newblock In \emph{ICCV}, 7919--7928.

\bibitem[{Yu et~al.(2021{\natexlab{b}})Yu, Rao, Zhao, Lu, and Zhou}]{yu2021groupawarecontrastiveregressionaction}
Yu, X.; Rao, Y.; Zhao, W.; Lu, J.; and Zhou, J. 2021{\natexlab{b}}.
\newblock Group-Aware Contrastive Regression for Action Quality Assessment.
\newblock In \emph{ICCV}, 7919--7928.

\bibitem[{Zhang et~al.(2025)Zhang, Li, Cheng, Hu, Yuan, Chen, Leng, Jiang, Zhang, Li, Jin, Zhang, Wang, Bing, and Zhao}]{zhang2025videollama3frontiermultimodal}
Zhang, B.; Li, K.; Cheng, Z.; Hu, Z.; Yuan, Y.; Chen, G.; Leng, S.; Jiang, Y.; Zhang, H.; Li, X.; Jin, P.; Zhang, W.; Wang, F.; Bing, L.; and Zhao, D. 2025.
\newblock VideoLLaMA 3: Frontier Multimodal Foundation Models for Image and Video Understanding.
\newblock arXiv:2501.13106.

\bibitem[{Zhang et~al.(2023)Zhang, Dai, Wang, Shen, Lu, Zhou, and Tang}]{zhang2024logolongformvideodataset}
Zhang, S.; Dai, W.; Wang, S.; Shen, X.; Lu, J.; Zhou, J.; and Tang, Y. 2023.
\newblock LOGO: A Long-Form Video Dataset for Group Action Quality Assessment.
\newblock In \emph{CVPR}, 2405--2414.

\bibitem[{Zhang et~al.(2024)Zhang, Zhang, Li, Zhao, Karypis, and Smola}]{zhang2023multimodal}
Zhang, Z.; Zhang, A.; Li, M.; Zhao, H.; Karypis, G.; and Smola, A. 2024.
\newblock Multimodal Chain-of-Thought Reasoning in Language Models.
\newblock arXiv:2302.00923.

\bibitem[{Zhou et~al.(2024{\natexlab{a}})Zhou, Cai, Wang, Shum, and Liang}]{zhou2024comprehensivesurveyactionquality}
Zhou, K.; Cai, R.; Wang, L.; Shum, H. P.~H.; and Liang, X. 2024{\natexlab{a}}.
\newblock A Comprehensive Survey of Action Quality Assessment: Method and Benchmark.
\newblock arXiv:2412.11149.

\bibitem[{Zhou et~al.(2024{\natexlab{b}})Zhou, Li, Cai, Wang, Zhang, and Liang}]{zhou2024cofinal}
Zhou, K.; Li, J.; Cai, R.; Wang, L.; Zhang, X.; and Liang, X. 2024{\natexlab{b}}.
\newblock CoFInAl: Enhancing Action Quality Assessment with Coarse-to-Fine Instruction Alignment.
\newblock In \emph{IJCAI}.

\bibitem[{Zhou et~al.(2025{\natexlab{a}})Zhou, Shum, Li, Zhang, and Liang}]{zhou2025phibridgingdomainshift}
Zhou, K.; Shum, H. P.~H.; Li, F. W.~B.; Zhang, X.; and Liang, X. 2025{\natexlab{a}}.
\newblock PHI: Bridging Domain Shift in Long-Term Action Quality Assessment via Progressive Hierarchical Instruction.
\newblock \emph{IEEE Trans. Image Process.}, 3718--3732.

\bibitem[{Zhou et~al.(2025{\natexlab{b}})Zhou, Zhu, Zhu, Wen, Liu, Xu, Meng, Cheng, Peng, Shen, and Feng}]{zhou2025chatvlaunifiedmultimodalunderstanding}
Zhou, Z.; Zhu, Y.; Zhu, M.; Wen, J.; Liu, N.; Xu, Z.; Meng, W.; Cheng, R.; Peng, Y.; Shen, C.; and Feng, F. 2025{\natexlab{b}}.
\newblock ChatVLA: Unified Multimodal Understanding and Robot Control with Vision-Language-Action Model.
\newblock arXiv:2502.14420.

\end{thebibliography}

\clearpage
\twocolumn[{%
\centering
\LARGE\bfseries HieroAction: Hierarchically Guided VLM for Fine-Grained Action Analysis\par
\vspace{6pt}
\Large Supplementary Material\par
\vspace{1em}
}]

\setlength{\leftmargini}{20pt}
\makeatletter
\def\@listi{\leftmargin\leftmargini \topsep .5em \parsep .5em \itemsep .5em}
\def\@listii{\leftmargin\leftmarginii \labelwidth\leftmarginii \advance\labelwidth-\labelsep \topsep .4em \parsep .4em \itemsep .4em}
\def\@listiii{\leftmargin\leftmarginiii \labelwidth\leftmarginiii \advance\labelwidth-\labelsep \topsep .4em \parsep .4em \itemsep .4em}
\makeatother

\setcounter{secnumdepth}{0}
\renewcommand\thesubsection{\arabic{subsection}}


\section*{Technical Details}

\subsection{Implementation Details}
We use Qwen2.5-VL-Instruct-7B~\cite{qwen2.5-VL} as the base model. Supervised fine-tuning (SFT) is conducted on four NVIDIA Tesla A100-40GB GPUs using the AdamW optimizer with a learning rate of $2 \times 10^{-5}$, a global batch size of 16 (per-device batch size 1 with accumulation steps of 4), and a random seed of 42. The model is trained for 25 epochs across three datasets: FineDive, FineFS, and LOGO.

In the second stage, HPL is performed using the same hardware and batch configuration as in SFT. The learning rate is set to $1 \times 10^{-6}$, and the KL divergence coefficient is $\beta = 0.04$. The model is trained for 10 epochs. For each input, $G = 8$ responses are sampled using a decoding temperature of 1.5 and a maximum output length of 3072 tokens. The response with the highest reward is selected for training.

\subsection{Evaluation Details}
Following the evaluation protocol in~\cite{chen2024motionllm}, we assess our method using GPT-4o. The evaluation returns a dictionary containing both the generated predictions and the corresponding score. The evaluation prompt used for this process is detailed in Table \ref{table:gpt-eval}.

\begin{figure*}[t]
\centering
\begin{tcolorbox}[
  enhanced,
  colback=white,
  colbacktitle=gray!40,
  coltitle=black,
  title={\textbf{Input:} question, answer, prediction},
  fonttitle=\bfseries,
  rounded corners,
  boxrule=0.5pt,
  arc=2mm,
  left=2mm, right=2mm, top=1mm, bottom=1mm,
  width=\textwidth
]

\textbf{LLM evaluation prompts:}

You are an intelligent chatbot designed for evaluating the correctness of generative outputs for question-answer pairs. Your task is to compare the predicted answer with the correct answer and determine if they match meaningfully. Here’s how you can accomplish the task:

\noindent\rule{\linewidth}{0.4pt}

\textbf{\#\#INSTRUCTIONS:}
\begin{itemize}
\item Check whether the reasoning process is clear, complete, and logically structured. Well-organized step-by-step explanations are preferred over vague or disjointed answers.
\item Determine if the predicted answer semantically aligns with the reference. Synonyms, rephrasings, or partial matches are acceptable as long as the core meaning is preserved.
\item Pay attention to the accuracy of action-related details (e.g., \texttt{sub\_actions}).
\end{itemize}

Please evaluate the following video-based question-answer pair: \\
\textbf{Question:} \{question\} \\
\textbf{Correct Answer:} \{answer\} \\
\textbf{Predicted Answer:} \{prediction\} \\

Provide your evaluation only as a yes/no and score where the score is an integer value between 0 and 5, with 5 indicating the highest meaningful match.\\
Please generate the response in the form of a Python dictionary string with keys \texttt{'pred'} and \texttt{'score'}, where the value of \texttt{'pred'} is a string of \texttt{'yes'} or \texttt{'no'} and the value of \texttt{'score'} is an \textbf{INTEGER}, not a string.\\
\textbf{DO NOT PROVIDE ANY OTHER OUTPUT TEXT OR EXPLANATION.} Only provide the Python dictionary string.\\

\textit{For example, your response should look like this: \texttt{\{'pred': 'yes', 'score': 4\}}.}

\end{tcolorbox}
\vspace{1mm}
\caption{\textbf{GPT evaluation prompt.} The prompt instructs GPT-4o to assess semantic alignment, reasoning completeness, and answer quality in a structured format.}
\label{table:gpt-eval}
\end{figure*}

\section*{Dataset Construction}

\begin{figure*}[t]
  \centering
  \includegraphics[width=\textwidth]{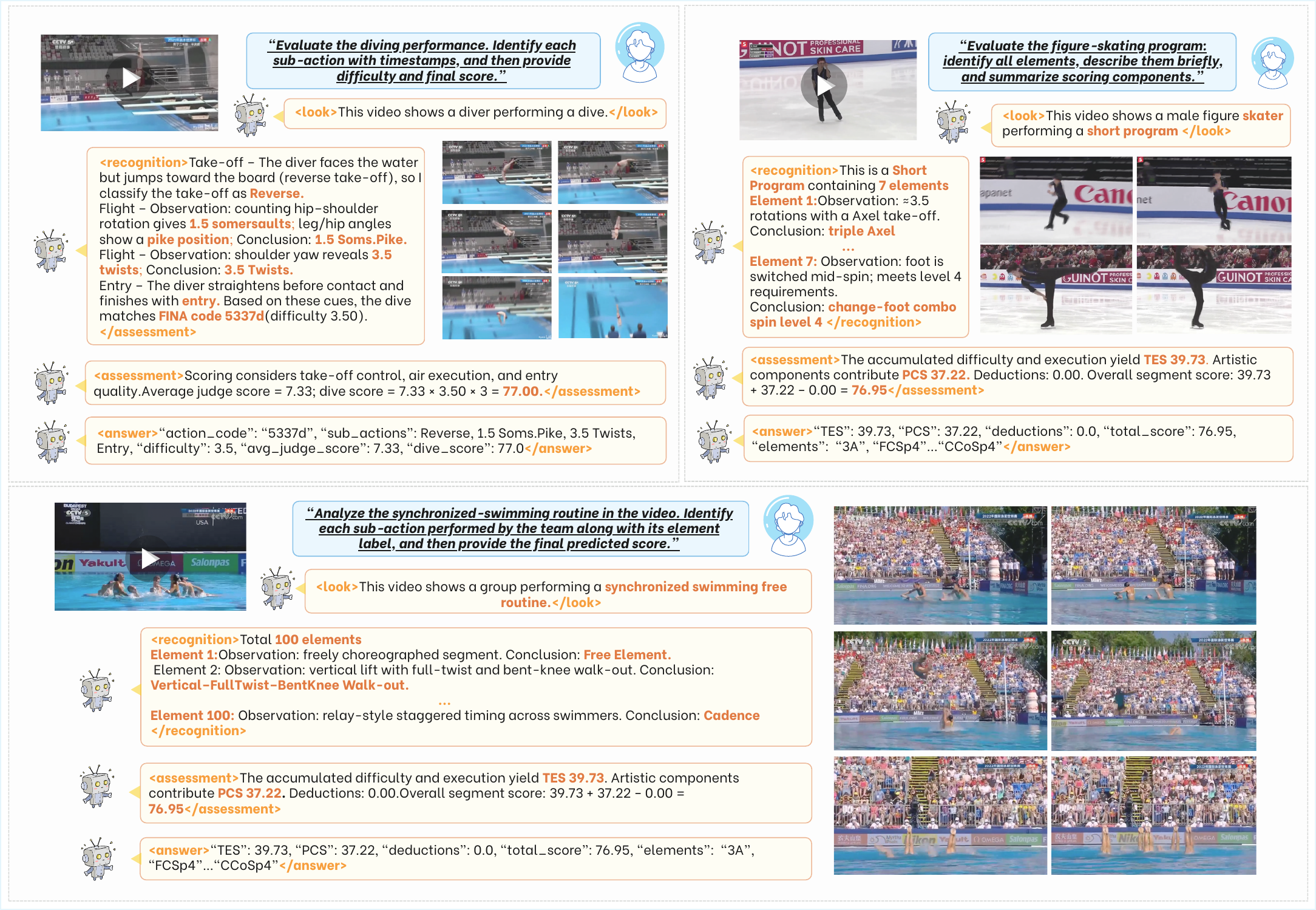}
  \caption{
Visualized reasoning outputs on the three sports datasets, illustrating the interpretable reasoning paradigm of our proposed framework.
}

  \label{fig:vis_supp}
\end{figure*}

This supplementary section provides a detailed overview of the datasets used in our study. To facilitate interpretable reasoning and expert-aligned evaluation, we construct question and answer (QA) datasets for fine-grained action understanding across three sports domains: diving, figure skating, and artistic swimming. These QA datasets are derived from existing fine-grained sports analysis datasets, which include action labels, sub-action annotations with time boundaries, and score information. Building on these resources, we reorganize the annotations into structured QA formats that support stage-wise reasoning across observation, recognition, and evaluation. The resulting datasets serve as the foundation for training and evaluating our proposed reasoning framework. We also provide visualizations of the model's reasoning process over these datasets to illustrate the interpretability and effectiveness of our structured approach, as shown in Figure~\ref{fig:vis_supp}.

\subsection*{FineDive}

FineDiveis~\cite{xu2022finedivingfinegraineddatasetprocedureaware} a high-quality, fine-grained dataset designed for competitive diving. It consists of 3000 short high-resolution video clips, each lasting around 10 to 20 seconds and covering the entire diving process from take-off to water entry. The dataset includes rich structured annotations, such as action labels, sub-action segments with time boundaries, difficulty coefficients, and judge scores. It features high temporal density and fine-grained class distinctions, covering a wide range of diving actions and difficulty levels. These properties make it a strong foundation for modeling both execution quality and evaluation logic.

Based on the natural structure and scoring rules of diving, we divide each dive into three key phases: take-off, flight, and entry. The take-off phase involves the direction and motion as the diver leaves the platform or springboard, providing the initial momentum for the dive. The flight phase includes rotations and somersaults performed in the air and is critical for assessing technical difficulty and precision. The entry phase captures the diver’s body position and splash control when entering the water, which significantly affects the final score. Notably, in 3-meter springboard events, the flight phase usually contains a single action, while in 10-meter platform dives, it may involve two consecutive sub-actions, making the structure more complex. To align with expert evaluation practices, we redesign the task structure based on the Stepwise Action Reasoning (SAR) framework. In the \textit{look} stage, the model identifies basic information about the diver—such as gender, stance, and platform type—to establish context. In the \textit{recognition} stage, original sub-action annotations are reorganized into three temporally aligned reasoning steps, each supported by paired observations and conclusions that guide the model from visual evidence to expert judgment. In the \textit{assessment} stage, the model simulates a judge’s scoring process for each phase and combines the sub-scores with the difficulty coefficient to produce a final score. As a result, FineDive is transformed into a structured, reasoning-oriented QA dataset. It enables the model to perform observation, recognition, and assessment in realistic sports scenarios, improving both interpretability and alignment with expert decision-making.

\subsection*{FineFS}

FineFS~\cite{JI2023FineFS} is a fine-grained figure skating dataset with 1,167 solo performance videos of 2–4 minutes each, including 933 for training and 234 for testing. Compared with short-duration sports such as diving, figure skating poses greater challenges in temporal localization and multi-dimensional scoring. Technical elements are sparsely distributed within continuous and highly artistic performances, making recognition and evaluation particularly demanding. The dataset provides detailed annotations—covering action types, sub-action segments, and both technical and program component scores—thus offering a solid foundation for addressing this complexity.

Building on this foundation, we design a QA task based on the SAR framework to support structured and interpretable scoring. Each performance follows one of two standard formats: the short program with seven required elements or the free skate with twelve. These elements include jumps, spins, and step sequences, all contributing to the final score. In the Look stage, the model identifies key information such as gender and program type, which helps estimate the number of elements and reduce misclassification. In the Recognition stage, the model detects and classifies each element sequentially, pairing it with visual observations and conclusion prompts to guide expert-like reasoning. In the Assessment stage, the model predicts both technical and program component scores and combines them into the final score, simulating figure skating’s dual scoring system. Training uses all score components for supervision, while evaluation focuses on the overall score. This design leverages the dataset’s annotations to improve recognition accuracy and better align predictions with expert judgments.

\subsection*{LOGO}

LOGO~\cite{zhang2024logolongformvideodataset} is a fine-grained artistic swimming dataset containing 200 team performance videos, with 150 for training and 50 for testing. Each video captures a complete 2–3 minute routine performed by a team of eight athletes. Compared with individual sports, artistic swimming is more complex for action understanding and score reasoning. Athletes must maintain precise synchronization, coordinated spatial formations, and consistent artistic expression, while the long temporal structure further complicates recognition and evaluation. By providing detailed annotations—including temporal boundaries of key choreographic segments and final scores from professional judges following FINA standards—the dataset lays a strong foundation for subsequent work aimed at these challenges.

Supported by this foundation, we design a QA task under the SAR framework to model the complex structure of artistic swimming routines and enable interpretable scoring. Each routine is classified as either a technical routine or a free routine, each with distinct rules and scoring criteria. In the Look stage, the model extracts contextual information such as the number of athletes, initial formations, and spatial layout, and identifies the routine type, providing critical context for later stages. In the Recognition stage, the model uses this context to infer the overall structure and analyze both temporal rhythm and spatial consistency among team members. For each segment, it identifies visual patterns and produces intermediate conclusions. In the Assessment stage, the model predicts scores based on FINA-aligned criteria, including synchronization, difficulty, and artistic impression. Evaluations on LOGO show the model’s ability to reason in long, multi-athlete performances while highlighting the challenges of interpretable team-based action analysis.

\end{document}